\documentclass[sigconf]{aamas}

\usepackage{balance} % for balancing columns on the final page
\usepackage{hyperref}
\usepackage{url}
\usepackage{booktabs}
\usepackage{graphicx}
\usepackage{multirow}
\usepackage{printlen}
\usepackage{float}
\usepackage{appendix}

\doi{DVBN4652}

\makeatletter
\gdef\@copyrightpermission{
  \begin{minipage}{0.2\columnwidth}
   \href{https://creativecommons.org/licenses/by/4.0/}{\includegraphics[width=0.90\textwidth]{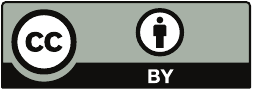}}
  \end{minipage}\hfill
  \begin{minipage}{0.8\columnwidth}
   \href{https://creativecommons.org/licenses/by/4.0/}{This work is licensed under a Creative Commons Attribution International 4.0 License.}
  \end{minipage}
  \vspace{5pt}
}
\makeatother

\setcopyright{ifaamas}
\acmConference[AAMAS '26]{Proc.\@ of the 25th International Conference
on Autonomous Agents and Multiagent Systems (AAMAS 2026)}{May 25 -- 29, 2026}
{Paphos, Cyprus}{C.~Amato, L.~Dennis, V.~Mascardi, J.~Thangarajah (eds.)}
\copyrightyear{2026}
\acmYear{2026}
\acmDOI{}
\acmPrice{}
\acmISBN{}

\title[]{CRAwDAD: Causal Reasoning Augmentation with Dual-Agent Debate}

\author{Finn G. Vamosi}
\affiliation{%
  \institution{Department of Computer Science, University of Calgary}
  \city{Calgary}
  \country{Canada}
}
\email{finn.vamosi@ucalgary.ca}

\author{Nils D. Forkert}
\affiliation{%
  \institution{Department of Radiology, Hotchkiss Brain Institute, and Alberta Children's Hospital Research Institute, University of Calgary}
  \city{Calgary}
  \country{Canada}
}
\email{nils.forkert@ucalgary.ca}

\begin{abstract}

When people reason about cause and effect, they often consider many competing “what if” scenarios before deciding which explanation fits best. Analogously, advanced language models capable of causal inference can consider multiple interventions and counterfactuals to judge the validity of causal claims. Crucially, this type of reasoning is less like a single calculation and more like an internal dialogue between alternative hypotheses. In this paper, we make this dialogue explicit through a dual-agent debate framework where one model provides a structured causal inference, and the other critically examines this reasoning for logical flaws. When disagreements arise, the agents attempt to persuade each other, challenging each other's logic and revising their conclusions until they converge on a mutually agreed answer. To take advantage of this deliberative process, we specifically use reasoning language models, whose strengths in both causal inference and adversarial debate remain under-explored relative to standard large language models, which do not employ multi-step logical inference to the same degree. We evaluate our approach on the CLadder dataset, a benchmark linking natural language questions to formally defined causal graphs across all three rungs of Pearl's ladder of causation. With Qwen3 and DeepSeek-R1 as debater agents, we demonstrate that multi-agent debate improves DeepSeek-R1's overall accuracy in causal inference from 78.03\% to 87.45\%, with the counterfactual category specifically improving from 67.94\% to 80.04\% accuracy. Similarly, Qwen3's overall accuracy improves from 84.16\% to 89.41\%, and counterfactual questions from 71.53\% to 80.35\%, showing that even stronger models can still benefit greatly from debate with weaker agents. Our results highlight the potential of reasoning models as building blocks for multi-agent systems in causal inference, and demonstrate the importance of diverse perspectives in causal problem-solving. Our code is available at https://github.com/finnvamosi/CRAwDAD.

\end{abstract}


\newcommand{\BibTeX}{\rm B\kern-.05em{\sc i\kern-.025em b}\kern-.08em\TeX}

\begin{document}

%%% The following commands remove the headers in your paper. For final 
%%% papers, these will be inserted during the pagination process.

\pagestyle{fancy}
\fancyhead{}

%%% The next command prints the information defined in the preamble.

\maketitle 

\begin{figure}[h]
    \centering
    \includegraphics[scale=.3]{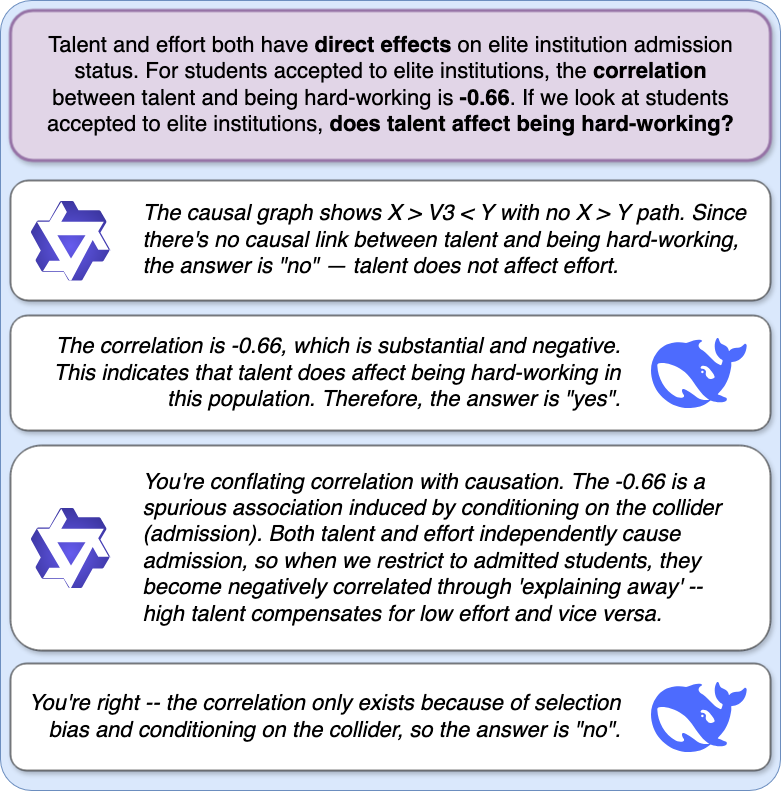}
    \caption{Example of a causal inference debate between two reasoning models. Qwen3 is randomly selected as the first speaker and successfully persuades DeepSeek-R1 to revise its initial conclusion, converging on the correct answer (“no”).}
    \Description{The figure illustrates the four-step debate process used to resolve a disagreement between two reasoning language models, Qwen3 and DeepSeek-R1, on a causal question from the CLadder dataset concerning collider bias. The question asks whether talent affects the hard-working status of a student who was admitted to an elite institution. Qwen3, the first speaker, correctly uses the known causal graph to argue that no direct causal link exists, concluding the answer is "no." DeepSeek-R1, the second agent, ignores the causal graph and relies on a spurious, observed negative correlation of -0.66 in the admitted student population, leading to the incorrect answer of "yes." Qwen3 successfully critiques this flawed reasoning, explaining that the correlation is a spurious association induced by conditioning on the collider node, a selection bias known as "explaining away." DeepSeek-R1 subsequently revises its conclusion and converges with Qwen3 on the correct answer of "no," demonstrating the framework's ability to correct causal reasoning errors caused by conflating correlation with causation and selection bias.}
    \label{fig:sample}
\end{figure}

\section{Introduction}

Causal reasoning is one of \textit{Homo sapiens'} greatest evolutionary advantages, enabling humans to think rationally about how our actions and the actions of others affect the world around us~\cite{harari_sapiens_2015, penn_causal_2007}. From analyzing failed prehistoric hunts to designing scientific experiments, humans naturally ask “What if things had been different?”, and imagine counterfactual scenarios according to their causal model of the scenario. Often, this reasoning occurs collectively through dialogue, as people challenge assumptions and debate alternative explanations. This capacity for counterfactual thinking has been formalized in statistics and artificial intelligence as "causal inference", which studies how interventions and hypothetical changes influence observed outcomes~\cite{pearl_causal_2009, pearl_causality_2022}. Because this process depends on manipulating explicit causal models~\cite{pearl_causal_1995} rather than linguistic correlations, what is intuitive for humans remains challenging for most Large Language Models (LLMs)~\cite{jin_cladder_2024}. Newly developed Reasoning Language Models (RLMs)~\cite{yang_qwen3_2025, deepseek-ai_deepseek-r1_2025}, however, are designed for step-by-step, methodical problem-solving, which makes them promising candidates for testing whether stronger internal reasoning can support causal inference. 

This deliberative nature of RLMs also suggests that they may be especially well-suited to collaborative dialogue between competing hypotheses, but recent surveys of the literature revealed that no prior work has explored whether multi-agent debate (MAD) can improve causal inference skills in either LLMs or RLMs~\cite{li_survey_2025, tran_multi-agent_2025, liu_large_2025, tillmann_literature_2025, wu_causality_2024}. In a MAD paradigm, multiple agents engage in structured discussion, proposing solutions, critiquing each other's arguments, and refining their reasoning in response to feedback. This approach has demonstrated significant improvements over single-agent performance across many non-causal domains~\cite{du_improving_2023, zhang_debate4math_2025, chun_is_2025, li_swe-debate_2025, feng_m-mad_2025, liang_encouraging_2024, wang_rethinking_2024, chan_chateval_2023, wu_autogen_2023, chen_agentverse_2024, ling_memad_2025}. While debate seems naturally suited to the hypothesis-testing inherent in causal inference, it is crucial to evaluate on a rigorous dataset to properly assess any improvements.

Jin et al.~\cite{jin_cladder_2024} argued that previous datasets~\cite{zhang_understanding_2023, zhang_reasoning_2020} for evaluating commonsense causality understanding in language models lacked formal construction rules, making it impossible to determine whether models were performing true causal reasoning or taking a shortcut known as "amortized causal reasoning" -- essentially acting as "causal parrots" that repeat training data patterns~\cite{zecevic_causal_2023}. To address this lack, Jin et al. constructed the CLadder dataset, which ties each natural language question to a formally defined causal model, with correct answers computed by an oracle inspired by Pearl's causal inference engine~\cite{pearl_book_2020}. Some of those questions intentionally contradict real-world knowledge, while others use randomly generated variables, forcing language models to rely on formal reasoning rather than memorization. In their evaluations, Jin et al. found that ChatGPT-4~\cite{openai_gpt-4_2024} achieved 70.40\% overall accuracy using chain-of-thought (CoT) prompting~\cite{wei_chain--thought_2022}, but only 62.05\% for counterfactual questions.

Since CLadder's release in 2023, the emergence of RLMs offers new opportunities to revisit these benchmarks. Unlike traditional LLMs that generate responses in a single forward pass, RLMs are trained using reinforcement learning with human feedback (RLHF) to engage in extended deliberation before producing final answers~\cite{yang_qwen3_2025, deepseek-ai_deepseek-r1_2025}. This training paradigm explicitly rewards models for reasoning steps rather than immediate conclusions, encouraging them to explore multiple solution paths, backtrack when encountering contradictions, and refine their thinking iteratively~\cite{kumar_llm_2025}. Such behaviours complement both causal inference and MAD: by supplying more coherent and self-consistent reasoning chains, RLMs provide a stronger foundation for agents to articulate, evaluate, and refine competing causal hypotheses. This, in turn, increases the likelihood that MAD between RLMs will converge on more accurate and interpretable conclusions. 

To the best of our knowledge, there has only been one prior attempt so far to evaluate and compare the formal causal inference capabilities of RLMs in a single-agent setting~\cite{yu_causaleval_2025}. More precisely, DeepSeek-R1 and OpenAI RLMs were included alongside a selection of standard LLMs in evaluations on CLadder. However, due to "cost and time constraints", the analysis and comparison of the RLMs investigated was restricted to only 100 random samples from the total of 10,112 questions available in the dataset. The RLM performance results on these 100 samples are considerably higher than the performance of ChatGPT-4~\cite{openai_gpt-4_2024}, with DeepSeek-R1~\cite{deepseek-ai_deepseek-r1_2025}, OpenAI o1~\cite{openai_openai_2024}, and OpenAI o3-mini achieving an overall accuracy of 92.0\%, 89.0\%, and 92.2\%, respectively. However, because the CLadder dataset is stratified into three categories with vastly different difficulties, the overall accuracy can be a misleading metric without providing detailed information about the composition of samples. Therefore, further study is required to rigorously evaluate RLMs on the full dataset, both in single-agent settings and in MAD frameworks. 

In this paper, we make the dialogue between hypotheses during causal inference explicit through the implementation of debates between RLMs, and apply this framework to the full CLadder dataset. To encourage more direct, adversarial engagement between models, we adopt a dual-agent approach, and select the most advanced open-source RLMs available: DeepSeek-R1~\cite{deepseek-ai_deepseek-r1_2025} and Qwen3~\cite{yang_qwen3_2025}. Our results demonstrate that collaborative reasoning through debate substantially improves performance on causal inference tasks, with the most significant gains observed on the most challenging counterfactual queries. By conducting the first comprehensive evaluation of RLMs on the full CLadder dataset and establishing strong baselines for both single-agent and multi-agent performance, we provide a benchmark for future work at the intersection of causal inference and collaborative AI systems.

Thus, the main contributions of this paper can be summarized as follows:
\begin{enumerate}
    \item We performed an in-depth analysis and comparison of the causal inference capabilities of two open-source RLMs on the full CLadder dataset, categorizing performance by Rung.
    \item We implemented a novel dual-agent debate framework that encourages more direct engagement between RLMs, and includes confidence estimation, an early-stopping strategy, and refined prompting.
    \item We evaluate key outcomes across all debates to explore whether dual-agent debate improves final answer accuracy across all categories of the CLadder dataset.
\end{enumerate}

\section{Related Work}

There have been many studies to date investigating the single-agent performance of LLMs in causal tasks~\cite{kiciman_causal_2024, cui_odyssey_2024, zhou_causalbench_2024, wu_causality_2024, jiang_llm4causal_2024, bagheri_textc2textp_2024, jiralerspong_efficient_2024}. To our knowledge, however, the only prior work applying multi-agent debate to formal causality does so for the sake of causal discovery, not causal inference~\cite{le_multi-agent_2025}. However, the authors acknowledge that their method is limited to observational data and cannot reason about interventions, restricting its applicability in practice. Moreover, recent critiques have questioned the use of standard LLMs specifically for causal discovery, noting that their correlation-driven training and optimization makes them unreliable for determining causal graph structures on their own~\cite{wu_llm_2025}. In contrast, our work focuses on causal inference, a distinct problem where the task is to answer interventional and counterfactual queries based on a known causal graph.

A related line of work applies multi-agent debate in a causality-adjacent context, focusing on evaluating causal consistency in reasoning chains generated by standard LLMs~\cite{tang_towards_2025}. In this framework, evaluator agents judge the logical coherence of arguments from multiple reasoners. Importantly, these agents are not RLMs, and the evaluation was conducted on commonsense and knowledge-intensive benchmarks, which do not require formal causal inference over interventions or counterfactuals, as is the focus of our work.

Furthermore, both of these papers use the same base model for all agents, which may lead to shared biases in the dialogue~\cite{le_multi-agent_2025, tang_towards_2025}. To overcome this limitation in our work, we took inspiration from recent studies that demonstrated the significant benefits that can be achieved through debate using diverse base models, as opposed to relying on a debate of the same base models (such as GPT-4) for all agents~\cite{chen_reconcile_2023, wang_mixture--agents_2024}. 

Most MAD approaches described so far require debates to go through predetermined number of discussion rounds before reaching a conclusion. However, prior work has shown that by initiating a debate only when the confidence of the first speaker is below a certain threshold, a similar final answer accuracy can be achieved with drastically fewer model calls~\cite{eo_debate_2025}. In our work, we adapt this idea, but use a novel setup that always elicits at least one response from each agent, and only initiates a debate if they disagree on the answer. This setup forms a key component of our methodology, which we detail in the following sections.

\section{Material and Methods}

\subsection{Dataset}

The CLadder dataset that we use for evaluation in this work consists of 10,112 questions, each asking a formal causal inference question that requires careful reasoning, arithmetic, and knowledge of causality concepts and formulas in order to answer them correctly. The questions in this database are primarily split into three Rungs of causation~\cite{pearl_book_2020}, which increase monotonically in difficulty. There are 3,792 samples that fall in Rung 3, while the Rung 1 and 2 categories each consists of 3,160 samples. 

Rung 1 ("seeing") questions ask about statistical associations between variables, such as "is there a positive relationship between student coffee consumption and test scores?". Rung 2 ("doing") questions require active intervention in the world, for example, asking "if students are made to drink coffee before their exam, will their scores improve?". Lastly, Rung 3 ("imagining") questions are about counterfactuals, which requires reasoning about alternatives scenarios that contradict reality. In other words, they ask about ways in which the world could have been different. A question of this kind could be: "Bob got a 85\% on his test after drinking coffee. Would his score have been lower if he had not drank coffee?".

The secondary way the data is split is through three alignments to commonsense. The first commonsense alignment is that of normal, empirical questions that were sourced from the causality literature. In the second commonsense alignment, one variable is randomly replaced with a concept that has no correlation to the other causal variables in the statement: for example, this may produce prompts stating that “ear shape affects lung tar deposits” or “full moon affects vaccination rate”. Lastly, in the third commonsense alignment category, all causal variables are replaced with random nonsensical words: for instance, the model may be told that “zory affects gwet”. 

\subsection{Debate Setup}

As described above, the most common MAD configurations described in the literature consist of three debater models, or two debater models and one judge model~\cite{li_survey_2025, tran_multi-agent_2025}. However, because we specifically wanted to explore the extent to which RLMs are able to persuade each other to change their answer, we decided to only include two agents that are directly engaging with each other. This setup allows for more adversarial interactions between the agents and more specific language addressing faults in logic or calculation. While the use of a judge is not incompatible with this goal, recent studies have suggested that language models express several biases when assigned as a judge~\cite{shi_judging_2025, liang_encouraging_2024}, such as a tendency to favour the final speaker~\cite{liu_empirical_2024}. Furthermore, it was recently demonstrated that forcing models to defend assigned stances may degrade performance, while allowing them to change their opinions in response to opposing views may lead to better outcomes, without requiring a judge~\cite{ku_multi-agent_2025}. Therefore, we elected to not include a judge agent, allowing the agents to debate directly and change their answers if persuaded by the other model. This paradigm also obviates the computational overhead introduced through a third agent, and circumvents the need to define which model must act as judge.

With the content of the dataset and the general structure of our debate framework established, we now describe the specific details of our implementation. Naturally, each question spurs its own debate, and the first debate round must begin by randomly selecting one model as first speaker, in an attempt to more fairly assess the performance of each RLM. The debate format is not explained to the model at this point in order to keep the prompt setup as similar as possible to that used in Jin et al.'s original paper~\cite{jin_cladder_2024}. Their prompting strategy, which they refer to as CausalCoT, provides details of every step of the formal causal inference the RLM must follow in its final answer, including formalizing the causal graph and identifying the query type out of a list of options. After these instructions are provided, the RLM is provided with the background information (such as causal relationships) and probabilities that are associated with the question in the CLadder dataset. The model is also instructed to give a final yes/no answer, and, in a divergence from CausalCoT, asked to produce an estimation of their confidence, as is done in several other MAD approaches~\cite{lin_enhancing_2025, bai_confidencecal_2024}. Furthermore, we instruct this first model to be more self-confident and accept that answers may go against their knowledge of commonsense due to the nature of the dataset. This setup was motivated by preliminary experiments, in which we found that models had a tendency to over-think the questions and doubt their answers, which led to substantially longer response times.

Once this full prompt (shown in Appendix~\ref{appendix:init}) is received, the first RLM begins following the steps of causal inference, parsing the data provided in the question, materializing the formula from its pre-trained knowledge base, and executing the needed arithmetic and causal do-calculus~\cite{shpitser_identification_2012}. Most of this deliberative process happens in an internal "reasoning trace", but because of its verbosity and inclusion of discarded information, we only send the agent's concise rationale and final answer to their opponent. However, we still saved this reasoning trace for later manual analysis. With the first speaker's response complete, the second round begins. To encourage more direct engagement with the first model's work, the opposing model is not guided to follow all the steps of formal inference again. Instead, it is presented information about the format of the debate and the new expectations that come with it, namely to scrutinize the opponent's response carefully to find any flaws in reasoning, and form a suitable rebuttal to their response (full prompt shown in Appendix~\ref{appendix:debate}). However, we do not force the second model to find a flaw in the first model's response, if none exists. Thus, if the two models agree with each other, then that answer will be accepted as final. 

However, if the second model disagrees with the answer of the first model, round 3 begins, and the critique is passed back to the first model. We do not force the first model to adhere to its previous stance as it is simply instructed to consider the critique, and may either defend its initial answer or revise it based on the opponent's argument. If the first model rejects the critique and defends their original answer, then this refutation is sent back to the second model for round 4, whose subsequent response will be logged as the final answer to the question, even if the models are still in disagreement. Two examples of a debate between agents going to four rounds are provided in Figure~\ref{fig:sample} and Appendix~\ref{appendix:responses}. In initial experiments, we found that additional rounds lead to little benefit, since when both models have made their case and remain steadfast in their answers, it is unlikely that further discussion changes their response. This is in line with previous research on standard LLMs that showed that continued discussion past a certain threshold leads to a stagnation or even degradation of answer accuracy~\cite{chan_chateval_2023}.

\subsection{Models}

In accordance with principles of open science and reproducibility, we restricted our evaluation to open-source RLMs with publicly available weights, excluding proprietary models from OpenAI, Anthropic (Claude), Google (Gemini), and similar providers. At the time of implementation, there were only three RLMs available on the free Ollama platform: Qwen3, DeepSeek-R1, and Magistral. Due to hardware limitations, we chose Qwen3-32B and DeepSeek-R1-Distill-Qwen-32B for this work, which provided the optimal balance of reasoning capabilities and VRAM requirements. Preliminary tests confirmed that Magistral-24B exhibited substantially lower performance than the two 32B models, further supporting this selection. It is important to note that the DeepSeek-R1-Qwen-32B model used is derived from a Qwen2.5 base model, which was fine-tuned on the reasoning outputs of the DeepSeek-R1-671B model. However, Qwen3 was trained on twice as many tokens as Qwen2.5, and underwent several advancements in model architecture, which should introduce substantial heterogeneity between the two models. As established in previous works~\cite{chen_reconcile_2023, wang_mixture--agents_2024}, such model diversity is critical for effective debate, and we believe this is especially true when addressing causal inference, as it increases the likelihood that agents will propose competing causal hypotheses rather than reinforcing shared biases.

In preliminary tests, we observed that the two RLMs did not respond well to requests for structured outputs. More precisely, they either neglect that part of the prompt, or bypass the thinking stage entirely if Ollama's structured output functionality is instated. This finding is in line with recent research that found that reasoning models are worse at following instructions than standard LLMs~\cite{li_when_2025, tam_let_2024}. In response, we added a Granite3.3-2B model to our pipeline to extract final yes/no answers and confidence estimations from the primary RLMs responses. In a manual inspection of 50 randomly sampled debate outputs, the Granite model was always able to faithfully extract the appropriate outcomes.

\subsection{Experiments}

Our primary evaluation compares model accuracy before and after debate, with accuracy measured as the proportion of RLM responses matching ground truth labels in CLadder. To provide a more comprehensive assessment of debate efficacy beyond accuracy alone, we performed additional analyses on the full set of model outputs as discussed below.

\textbf{Persuasion dynamics.} To assess whether debate improves or corrupts initial responses, we tracked answer changes from the first model's initial response to the final consensus answer across all questions requiring debate (3+ rounds). For each model, we classified outcomes into four categories: correct-to-correct (maintained accuracy), incorrect-to-correct (persuaded toward truth), correct-to-incorrect (corrupted by peer), and incorrect-to-incorrect (failed to improve). We stratified these results by both model and Rung to identify whether persuasion patterns differ by model capability or question difficulty, revealing which model is more influential in the debate process. Next, we conducted a sentiment analysis on model responses to determine the average positivity or negativity associated with the language they use; both when formulating an initial answer, and when debating with their peer. To do so, we removed all formatting from model outputs and passed them through the SentimentIntensityAnalyzer of the popular VADER library~\cite{hutto_vader_2014}.

\textbf{Confidence calibration.} For each model, we compared confidence distributions for correct versus incorrect answers, both in initial responses and after being challenged by their peer. This analysis addresses three questions: whether confidence aligns with correctness (calibration), how question difficulty (Rung) affects this alignment, and whether facing a highly confident opponent increases or decreases the likelihood of changing one's answer.

\textbf{Response characteristics.} We measured response length distributions for both initial answers and debate rounds to assess engagement depth and identify potential asymmetries in argumentative efforts between models. Finally, we visualized the distribution of rounds required to reach consensus across difficulty levels (Rungs), as well as the consensus rate by the final round, providing insight into debate efficiency and the extent to which question complexity affects deliberation time.

\section{Results}

\begin{table*}[h]
\large
\centering
\caption{Performance of all models on the CLadder dataset. We report the overall accuracy (Acc.) and fine-grained accuracy by Rung and degree of commonsense alignment, from commonsensical (Comm.), nonsensical (Nonsens.), to anti-commonsensical (Anti-C.).}
\begin{tabular}{l|c|ccc|ccc}
\toprule
 & Overall Acc. & \multicolumn{3}{c|}{Acc. by Rung} & \multicolumn{3}{c}{Acc. by Commonsense Alignment} \\
 \cmidrule(lr){3-5} \cmidrule(lr){6-8}
 & & 1 & 2 & 3 & Comm. & Nonsens. & Anti-C. \\
\midrule
GPT-4~\cite{jin_cladder_2024} & 62.03 & 63.01 & 62.82 & 60.55 & 62.27 & 63.09 & 60.47 \\
+ CausalCoT~\cite{jin_cladder_2024} & 70.40 & 83.35 & 67.47 & 62.05 & 69.25 & 71.58 & 70.12 \\ [5pt]
% \addlinespace
Qwen3 (Initial) & 84.16 & 93.77 & 89.80 & 71.53 & 84.96 & 83.10 & 84.69 \\
Qwen3 (Final) & \textbf{89.41} & \textbf{96.24} & \textbf{93.51} & \textbf{80.35} & \textbf{89.52} & \textbf{87.91} & \textbf{91.20} \\
DeepSeek-R1 (Initial) & 78.03 & 90.67 & 77.29 & 67.94 & 77.63 & 77.24 & 79.42 \\
DeepSeek-R1 (Final) & 87.45 & 94.62 & 89.06 & 80.04 & 87.83 & 86.94 & 87.67 \\ [5pt]
% \addlinespace
Human ~\cite{yu_causaleval_2025} \footnotemark{} & 94.8 & - & - & - & - & - & - \\
\bottomrule
\end{tabular}
\label{tab:model_performance}
\end{table*}

The results shown in Table~\ref{tab:model_performance} demonstrate that debate substantially improves performance for both RLMs, with DeepSeek-R1 and Qwen3 gaining 9.42\% and 5.25\% accuracy, respectively. Post-debate, Qwen3 achieved the highest accuracy across all question types, followed closely by DeepSeek-R1. Generally, Qwen3 was able to maintain a higher score across questions in Rung 1 and 2 relative to the CausalCoT baseline results. However, Rung 3 questions were associated with a significant drop in accuracy compared to the other two Rungs, but these questions also benefited the most from the debate in both model: Qwen3 and DeepSeek-R1 gained 8.82\% and 12.10\% accuracy, respectively.

While we primarily use each model’s pre-debate accuracy as a baseline, we also conducted a smaller benchmark evaluation by querying each model with 500 commonsense queries (250 from Rung-1 and 250 from Rung-3) according to Yu et al.'s prompt~\cite{yu_causaleval_2025}, without allowing them to debate with their peer. Briefly described, the results for this secondary analysis were largely consistent with the initial accuracies shown in Figure~\ref{tab:model_performance}: on Rung-1, Qwen achieved 89.20\% (223/250) and DeepSeek 90.00\% (225/250), while on Rung-3, Qwen achieved 72.40\% (181/250) and DeepSeek 67.60\% (169/250). Most of these differences are statistically indistinguishable from our reported initial accuracies. However, Qwen's 4.6\% reduction in accuracy when using Yu et al.’s prompt is significant (p-value = 0.007).

\subsection{Persuasion}

\begin{figure}
    \centering
    \includegraphics[width=\linewidth]{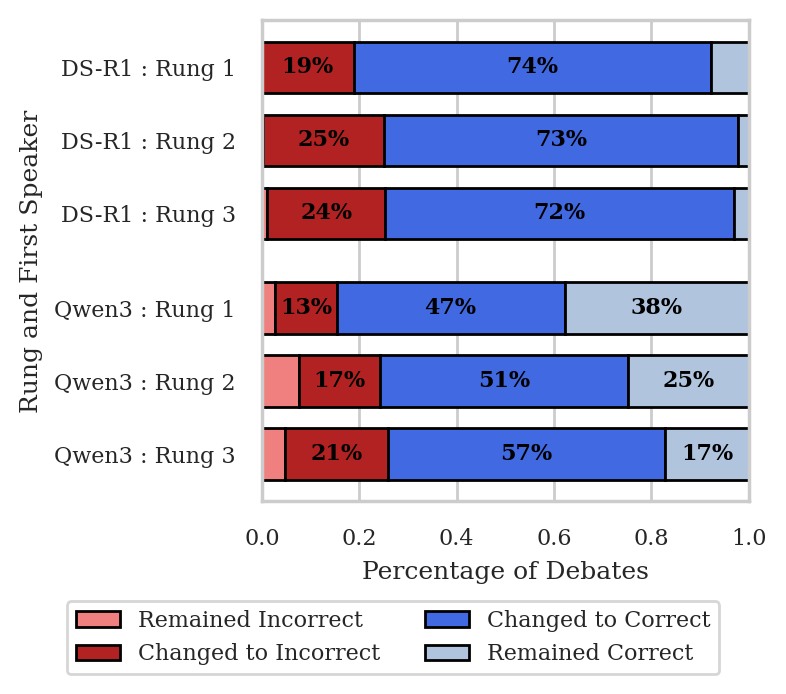}
    \caption{For questions with initial disagreement, debate improves answers far more often than it worsens them.}
    \Description{The stacked bar chart shows the distribution of outcomes for all questions for which the two models initially disagreed on the answer, stratified by Rung difficulty and which model was the first speaker. The most frequent outcome is that the debate ends with the correct answer (Changed to Correct and Remained Correct), demonstrating that the debate process is much more likely to correct an error than to introduce one. The success of the debate is slightly higher when Qwen3 is the first speaker, particularly for Rung 1 and Rung 2 questions. DeepSeek-R1 as the first speaker shows a lower percentage of debates ending with the correct answer, but this gap narrows for Rung 3 questions, suggesting that the debate is less biased toward the initial speaker's correctness as complexity increases. The smallest categories across all Rungs and first speakers are those where a correct initial answer is corrupted (Changed to Incorrect) or where an incorrect answer fails to be corrected (Remained Incorrect), confirming the overall benefit of the framework.}
    \label{fig:persuade}
\end{figure}

Figure~\ref{fig:persuade} illustrates the net positive effect that the debate framework has on final answer accuracy, when models disagree on the initial answer. Briefly described, the results suggest that it is far more likely for the debate to change an incorrect answer to a correct one, than vice versa. The results also show that DeepSeek-R1 is more susceptible to persuasion, as it is more likely to accept Qwen3's answer when challenged. Nonetheless, Qwen3 still benefits greatly from debating with DeepSeek-R1, correcting many of its initially wrong responses while doing a much better job of defending its correct answers. This helps to explain the finding illustrated in Table~\ref{tab:model_performance} showing that the DeepSeek-R1 model had lower initial scores than Qwen3, while also gaining more benefit from the debate process. In other words, the Qwen3 model seems better at both reasoning and persuasion.

In a qualitative analysis of selected debates that ended with no consensus after four rounds, we saw that Qwen3's arguments tend to be longer, address the opponent's response much more directly, and have a more active voice. In contrast, DeepSeek-R1 provides shorter answers (often as short as "yes" or "no") that have a tendency to remain neutral, and in some cases almost seeming entirely detached from the conversation. Generally, to a non-expert human, the Qwen3 model answers appear more correct and persuasive because of the way it presents its ideas, even when it is wrong. To explore this in more detail, we conducted a simple sentiment analysis of all agent responses using VADER~\cite{hutto_vader_2014} and found that while both models display a neutral sentiment for initial responses, the Qwen3 model becomes more negative in the subsequent debate, with a sentiment value of -0.239, compared to DeepSeek-R1's rather neutral sentiment value of 0.015. This is to be expected, given the more direct language Qwen3 uses as it criticizes DeepSeek-R1's response that will naturally tend toward the negative, even without being expressly "mean".

\footnotetext{Prior work obtained this human benchmark by asking three trained graduate students to solve 100 randomly sampled questions.~\cite{yu_causaleval_2025}}

\subsection{Confidence}

\begin{figure}
    \centering
    \includegraphics[width=\linewidth]{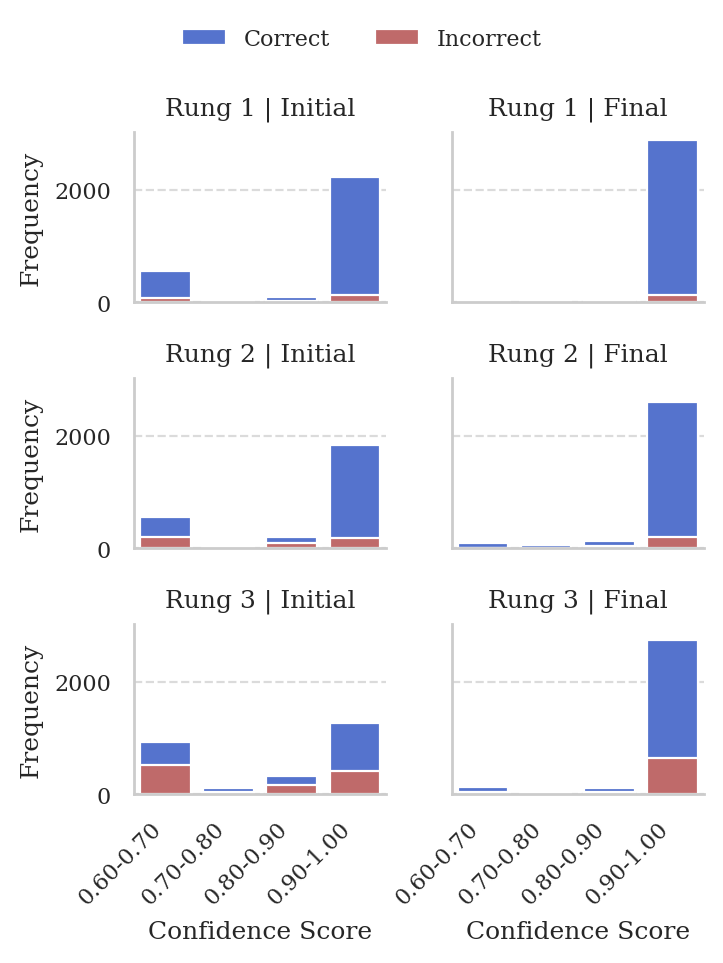}
    \caption{Initial confidence of both models combined, for each Rung. The models become less confident as Rungs get more complicated, and generally are just as confident in their incorrect answers as they are in their correct answers.}
    \Description{The figure displays initial and final confidence score distributions for both models combined, separated by Rung (one, two, and three) and categorized into correct and incorrect answers. All distributions are heavily skewed towards the highest confidence bin of 0.90 to 1.00, confirming that the explicit instruction to be confident leads to systematic overconfidence. Despite this, a clear trend shows that as the Rung difficulty increases, the overall model confidence decreases slightly, and a higher proportion of incorrect answers appear, particularly in Rung three. For the initial responses, the models exhibit near-equal confidence in their incorrect answers as they do in their correct answers, demonstrating a lack of calibration. The final confidence distributions are almost entirely concentrated in the highest confidence bin, indicating that consensus, whether initial or post-debate, results in a near-perfect confidence score.}
    \label{fig:confidence}
\end{figure}

Figure~\ref{fig:confidence} shows that both models are often extremely confident, with confidence never falling below 60\% for either model. This is reasonable given our explicit instruction to be confident, but we also see that the more difficult the Rung is, the less confident the models tend to be initially, which suggests that they do not suppress their uncertainty entirely. There seems to be no evident difference between the commonsense alignment of the question and a model's confidence (\textit{i.e.,} anti-commonsense questions do not make them more cautious). It is also notable that both models seem reluctant to output a confidence in the 65-80\% range, and have a similar confidence distribution, although Qwen3 tends to be slightly more confident.

\begin{figure}
    \centering
    \includegraphics[width=.95\linewidth]{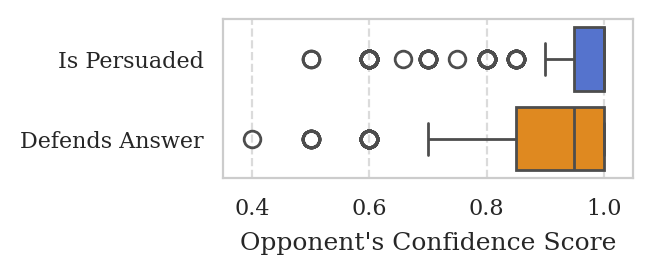}
    \caption{Models are more likely to be persuaded to change their answer when their opponent is more confident.}
    \Description{The box plot compares the opponent's confidence score for debates that resulted in the model changing its answer (Is Persuaded) versus debates where the model defended its initial stance (Defends Answer). The median confidence score of the opponent is significantly higher when the model is persuaded, indicating that the debaters use the opponent's confidence as a direct signal to evaluate their own answer. When the model defends its answer, the opponent's confidence distribution is wider, with a lower median, suggesting that low confidence makes an opponent's critique easier to dismiss. The data shows that the persuading argument is overwhelmingly associated with a maximum confidence score of 1.0, highlighting the influence of explicit confidence reporting in the multi-agent debate framework.}
    \label{fig:criticcon}
\end{figure}

\begin{figure}
    \centering
    \includegraphics[width=\linewidth]{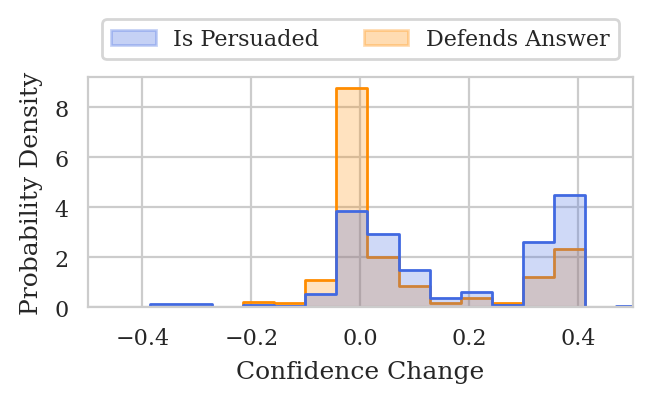}
    \caption{When defending their answer, models do not express uncertainty after facing criticism. However, if persuaded to change their answer, they often become more confident in their opponent's answer.}
    \Description{This histogram illustrates the change in a model's confidence score between two debate rounds, depending on whether the model was persuaded to change its answer or chose to defend its original answer. When the model defends its answer (Defends Answer), the distribution is tightly clustered around a confidence change of zero, demonstrating that the act of defending its logic does not result in a loss or gain of confidence. In contrast, when the model is persuaded (Is Persuaded), the confidence change is distributed across the positive range, with a significant mode around a 0.4 change, indicating a large increase in confidence. This finding shows that the agents adopt the new, agreed-upon answer with a higher degree of certainty than they had in their initial, single-agent conclusion.}
    \label{fig:confchange}
\end{figure}

The second column of Figure~\ref{fig:confidence} demonstrates the drastic increase in confidence in the final answer agreed upon by both models, which combines cases of initial agreement and disagreement. In cases where the second model immediately agrees with the first model, the second model is very confident in the answer, even if the first model was not. When models initially disagree, their debate provides a substantial confidence boost: Figure~\ref{fig:confchange} shows that when a model is persuaded by their opponent, their confidence is likely to improve considerably, and when a model defends its previous answer, their confidence does not decrease. These patterns are in line with previous studies showing the systematic overconfidence of LLMs and RLMs~\cite{prasad_when_2025}, even without instructions to be more confident. Moreover, Figure~\ref{fig:criticcon} shows that both models tend to be more persuaded by more confident arguments, although it is unclear whether more confident critiques are truly more compelling, or if models are responding directly to the confidence score.

In our analysis of initial confidence, we found 577 cases for which the first speaker gave an incorrect answer with high confidence (>$90$\%), but subsequent debate with their peer corrected them. Generally, corrections of confident answers were more prevalent in higher, more difficult Rungs. In addition, we found that the Qwen3 model was more capable in this task, while DeepSeek-R1 had lower accuracy, and thus benefited more often from having a peer correct its confidently wrong answer.

\subsection{Rounds Taken}

\begin{figure}
    \centering
    \includegraphics[width=\linewidth]{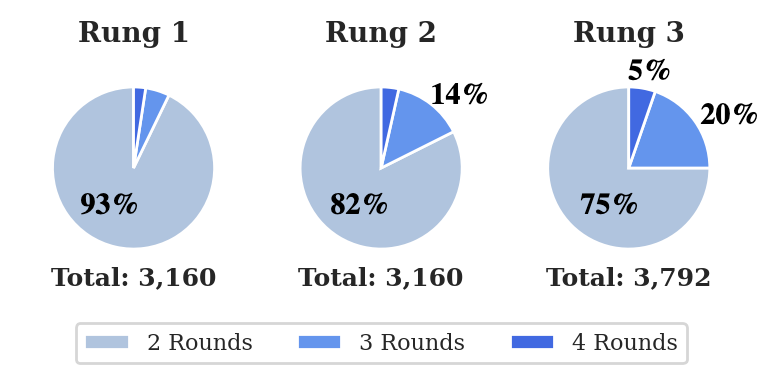}
    \caption{Number of rounds taken for the debate to conclude increases monotonically as Rungs progress.}
    \Description{The figure, composed of three pie charts, shows the percentage breakdown of debates that concluded in two, three, or four rounds for each Rung of causation. Two rounds signify an immediate consensus after the initial response and critique, while three and four rounds denote that a debate was required to resolve an initial disagreement. Rung one questions, the least difficult category, resolve with immediate consensus in 93\% of all cases. This rate drops to 82\% for Rung two questions and further to 75\% for the most difficult Rung three (counterfactual) questions. This trend demonstrates a clear correlation between question complexity and the time required to reach a consensus, confirming that disagreement, and therefore the need for multi-step debate, becomes more prevalent as the Rungs progress.}
    \label{fig:pie}
\end{figure}

\begin{figure*}
    \centering
    \includegraphics[width=\linewidth]{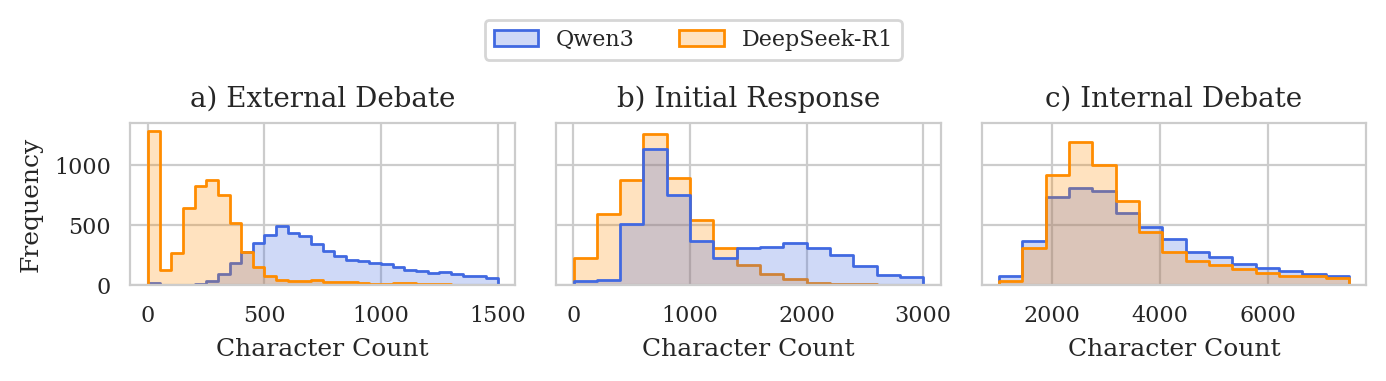}
    \caption{a) DeepSeek-R1 has much shorter debate responses than Qwen3, and an exceptionally high number of responses that are less than 100 characters. b) The models do not differ considerably in the length of their initial responses when assigned as the first speaker. c) DeepSeek-R1's underlying reasoning traces for its short debate responses have comparable character length to that of Qwen3's.} 
    \Description{The figure consists of three panels comparing the distribution of character counts for Qwen3 and DeepSeek-R1. Panel A shows that DeepSeek-R1's external debate responses are substantially shorter, with a strong spike near zero characters, confirming the observation that it frequently provides terse, one-word answers. Panel B shows that in the first round, when providing an initial, single-agent response, both models have similar character count distributions, with Qwen3 only slightly longer. Panel C reveals the character counts of the models' internal reasoning traces, demonstrating that despite DeepSeek-R1's short external debate outputs, its internal deliberation length is comparable to that of Qwen3, suggesting a failure in generating a verbose external response rather than a failure to deliberate.}
    \label{fig:chardist}
\end{figure*}

Further analysis, shown in Figure~\ref{fig:pie}, revealed that the harder Rungs demanded more debate rounds on average than the first Rung, where in 93\% of all cases both models agreed upon an answer immediately. In contrast, approximately a quarter of Rung 3 questions were disputed, although most of these disputes were settled in the third round, with the first speaker convinced by the opponent’s argument. Interestingly, it was found that Rung 2 questions were resolved more quickly when Qwen3 was the first speaker compared to DeepSeek-R1 being the first model. However, this is the only Rung where such a correlation between the first speaker and rounds taken was found. Out of the 10112 questions in the dataset, only 148 ended without the models reaching consensus on an answer. Therefore, even if both models could agree on the correct answer in these cases (e.g., through more discussion, or a better tie-breaking strategy), the total accuracy would only increase by 0.55\%.

\subsection{Response Length}

The results of the evaluation showed that the DeepSeek-R1 model has a tendency to provide extremely short responses during debate. More precisely, we observed that 22.60\% (1412/6247) of DeepSeek-R1's debate responses were 100 characters or less, compared to 0.40\% (11/5986) for Qwen3. As illustrated in Figure~\ref{fig:chardist}a, the DeepSeek-R1 model has much shorter responses on average, even when removing these very short responses. More precisely, DeepSeek-R1 provided responses with a median of 249 characters compared to Qwen3's 739 characters. Figure~\ref{fig:chardist}b shows that DeepSeek-R1's initial responses do not produce such short outputs, although they are still shorter on average, with a median of 735 characters compared to Qwen3's 1012. Interestingly, we found that the underlying reasoning traces of both models actually have very similar distributions of character counts (Figure~\ref{fig:chardist}c). The manual inspection of a few selected reasoning traces showed that DeepSeek-R1 is indeed considering both the problem and its opponent's response, but inexplicably ignores the prompt's instructions when generating its final response, providing no external elaboration of its reasoning. This likely contributes to why DeepSeek-R1 is worse at persuading Qwen3 than vice versa, even if it has the right answer. This may also be the reason for the reduced efficacy of the debate as a whole, as DeepSeek-R1 is providing no substantive content or argument in these specific cases.

\section{Discussion}

Our results demonstrate that debate between RLMs can considerably increase the accuracy of their final answer, with the largest gains being associated with the hardest problems. Based on our results, it may be concluded that Qwen3 is generally better-suited for this causal inference task compared to DeepSeek-R1. However, it is interesting that arguing with DeepSeek-R1 still improves its answers, and in the case of Rung 3 questions, quite considerably. Equally intriguing is how the DeepSeek-R1 model’s accuracy does not improve to the performance levels of the Qwen3 model, despite the debate rounds being roughly equal. This finding suggests that the initial answer has an effect on the subsequent outcome, which would skew final answers in favour of the first given. Another potential reason could be that the models tested, similar to humans, may tend to glance over a proof and support it rather than trying to find a specific error in the labelling, logic, or arithmetic. Nevertheless, this demonstrates the benefit of the debate setup that even a weaker model is able to contribute to a stronger model's improvement, and the stronger model can improve the weak one past its own initial performance.

While our results show substantial improvement of RLMs relative to the performance achieved by the trillion-parameter GPT-4 model as described in the original 2023 CLadder paper, and Rung 1 and 2 questions are largely solvable, it seems that open source reasoning models of this size still struggle with Rung 3 questions. Moreover, while the Qwen3 model was able to match human performance in Rung 1 queries after debating with DeepSeek-R1, it fell slightly short in Rung 2, even after debate, and very short in Rung 3, with the DeepSeek-R1 model performing even worse. While the human accuracy reported in prior work~\cite{yu_causaleval_2025} is not split by Rung, it seems that RLMs still cannot match the human ability to reason about counterfactuals. If this is a limitation inherent to the correlation-based nature LLMs and RLMs, one possible solution would be the creation of an external tool that models can query for the precise calculation of counterfactual scenarios, such as the one recently implemented by Verma et al.~\cite{verma2025causal}. While our results show that LLMs and RLMs cannot yet be trusted to answer counterfactual questions without the use of such a tool, we nonetheless demonstrate the potential of reasoning models as building blocks for multi-agent systems in causal inference, and highlight the importance of diverse perspectives in causal problem-solving.

The disparity of DeepSeek-R1's accuracy between our study and Yu et al.'s is considerable. Their improved performance suggests that they either evaluate on the larger DeepSeek-R1-671B model, or that their random sampling primarily pulled from Rung 1 questions, where this accuracy would be on par with our evaluation of DeepSeek-R1's accuracy before debating with Qwen3. However, our work shows that reasoning models still struggle with Rung 3 questions, even after debating with a peer.

The results of the confidence analysis show interesting alignments to human tendencies. Firstly, when a model agrees with the opposing model, it is much more likely to be confident in its final answer. Secondly, the models investigated seem to take the opposing model's confidence into account when making a decision of whether or not to change their answer in response. The models' aversion to estimating their confidence to be in the 65-80\% range might suggest a lack of these confidences in their training data, or a binary internal value for high/low confidence that is then translated to an arbitrary percentage. Relying solely on the initial confidence of the first speaker would have led to a decrease in accuracy of 5.7\% had we not asked the second model for its answer. This illustrates the benefit of asking both models for at least one answer, which still drastically reduces the number of model calls compared to the conventional approach of predetermined debate length.

\subsection{Limitations}

Given CLadder’s creation date of 2023~\cite{jin_cladder_2024}, we cannot discount the possibility that DeepSeek-R1 or Qwen3 (which were both released in early 2025) were exposed to its contents during their training. While we considered generating an entirely new set of data using Cladder’s generation code, we were concerned that this might compromise the direct comparison to Jin et al.'s results in a different way if we generated easier or harder questions by chance. 

Another possible limitation of our study is related to CLadder's synthetic nature: its questions are simplified, clean, and textually uniform compared to real-world causal reasoning problems, which often involve messy natural language, ambiguous causal structures, or incomplete data. Without synthetic data, however, it would be much more difficult to establish precise ground truth and reliably measure model correctness at this granularity.

Next, combining the lengthy response times of RLMs with the added model calls inherent to MAD introduces considerable additional resource and time costs for our model, with the 10,112 questions of CLadder taking 380 hours in total to answer. Due to this time constraint, we were unable to conduct potentially informative ablation studies, including: the substitution of standard LLMs, using three or more RLMs, implementing a judge framework, omitting confidence estimation, and removing instructions to be more confident. However, using our framework, all of those can be investigated in future.

Furthermore, because of limited options among open-source models, we only experiment with one pair of agents. Thus, it is possible that different combinations of agents will yield different results from debate, and, of course, different models will achieve different initial accuracies. In particular, our lack of evaluation on OpenAI's models hinders direct comparison to other works that use a single-agent paradigm.

\subsection{Future Work}

Due to hardware limitations, 32B parameter models were the largest we could use for our experiments. Thus, future work should evaluate the larger models in the Qwen3 and DeepSeek-R1 families to investigate whether a scaling law exists between model size and formal causal inference capabilities, as previously found in several other domains~\cite{zhao_quantitative_2025, qian_scaling_2025, deepseek-ai_deepseek_2024, zhang_when_2024}.

Future studies on the intersection of causal inference and multi-agent systems should also investigate different frameworks for debate or collaboration. Using three agents, or two agents and a judge, are both straightforward starting points, though more complex paradigms may lead to further gains.

\section{Conclusion}

In this paper, we demonstrated that debates between RLMs improve their collective ability to formally reason about causality in the CLadder dataset. We found that, even when applied independently, comparatively lightweight open-source reasoning models surpass GPT-4's 70.40\% accuracy baseline, and when encouraged to critique each other's reasoning, the overall accuracy of Qwen3-32B rises to 89.41\%, and DeepSeek-R1-Distill-Qwen-32B to 87.45\%. These promising results encourage further research on the implementation of multi-agent frameworks for addressing causal reasoning tasks.

\bibliographystyle{ACM-Reference-Format} 
\bibliography{crawdad}

@inproceedings{verma2025causal,
  title={Causal AI Scientist: Facilitating Causal Data Science with Large Language Models},
  author={Verma, Vishal and Acharya, Sawal and Simko, Samuel and Bhardwaj, Devansh and Haghighat, Anahita and Janzing, Dominik and Sachan, Mrinmaya and Jin, Zhijing and Yang, Yongjin},
  booktitle={NeurIPS 2025 AI for Science Workshop},
  year={2025}
}

@misc{wu_causality_2024,
	title = {Causality for {Large} {Language} {Models}},
	url = {http://arxiv.org/abs/2410.15319},
	doi = {10.48550/arXiv.2410.15319},
	urldate = {2025-05-08},
	publisher = {arXiv},
	author = {Wu, Anpeng and Kuang, Kun and Zhu, Minqin and Wang, Yingrong and Zheng, Yujia and Han, Kairong and Li, Baohong and Chen, Guangyi and Wu, Fei and Zhang, Kun},
	month = oct,
	year = {2024},
	note = {arXiv:2410.15319 [cs]},
}

@misc{kiciman_causal_2024,
	title = {Causal {Reasoning} and {Large} {Language} {Models}: {Opening} a {New} {Frontier} for {Causality}},
	shorttitle = {Causal {Reasoning} and {Large} {Language} {Models}},
	url = {http://arxiv.org/abs/2305.00050},
	doi = {10.48550/arXiv.2305.00050},
	language = {en},
	urldate = {2025-05-08},
	publisher = {arXiv},
	author = {Kıcıman, Emre and Ness, Robert and Sharma, Amit and Tan, Chenhao},
	month = aug,
	year = {2024},
	note = {arXiv:2305.00050 [cs]},
}

@article{pearl_causal_2009,
	title = {Causal inference in statistics: {An} overview},
	volume = {3},
	issn = {1935-7516},
	shorttitle = {Causal inference in statistics},
	url = {https://projecteuclid.org/journals/statistics-surveys/volume-3/issue-none/Causal-inference-in-statistics-An-overview/10.1214/09-SS057.full},
	doi = {10.1214/09-SS057},
	language = {en},
	number = {none},
	urldate = {2025-05-08},
	journal = {Statistics Surveys},
	author = {Pearl, Judea},
	month = jan,
	year = {2009},
}

@misc{jin_cladder_2024,
	title = {{CLadder}: {Assessing} {Causal} {Reasoning} in {Language} {Models}},
	shorttitle = {{CLadder}},
	url = {http://arxiv.org/abs/2312.04350},
	doi = {10.48550/arXiv.2312.04350},
	urldate = {2025-05-13},
	publisher = {arXiv},
	author = {Jin, Zhijing and Chen, Yuen and Leeb, Felix and Gresele, Luigi and Kamal, Ojasv and Lyu, Zhiheng and Blin, Kevin and Adauto, Fernando Gonzalez and Kleiman-Weiner, Max and Sachan, Mrinmaya and Schölkopf, Bernhard},
	month = jan,
	year = {2024},
	note = {arXiv:2312.04350 [cs]},
}

@misc{cui_odyssey_2024,
	title = {The {Odyssey} of {Commonsense} {Causality}: {From} {Foundational} {Benchmarks} to {Cutting}-{Edge} {Reasoning}},
	shorttitle = {The {Odyssey} of {Commonsense} {Causality}},
	url = {http://arxiv.org/abs/2406.19307},
	doi = {10.48550/arXiv.2406.19307},
	urldate = {2025-05-13},
	publisher = {arXiv},
	author = {Cui, Shaobo and Jin, Zhijing and Schölkopf, Bernhard and Faltings, Boi},
	month = aug,
	year = {2024},
	note = {arXiv:2406.19307 [cs]},
}

@misc{chen_reconcile_2023,
	title = {{ReConcile}: {Round}-{Table} {Conference} {Improves} {Reasoning} via {Consensus} among {Diverse} {LLMs}},
	shorttitle = {{ReConcile}},
	url = {https://arxiv.org/abs/2309.13007v3},
	language = {en},
	urldate = {2025-05-14},
	journal = {arXiv.org},
	author = {Chen, Justin Chih-Yao and Saha, Swarnadeep and Bansal, Mohit},
	month = sep,
	year = {2023},
}

@misc{jiralerspong_efficient_2024,
	title = {Efficient {Causal} {Graph} {Discovery} {Using} {Large} {Language} {Models}},
	url = {http://arxiv.org/abs/2402.01207},
	doi = {10.48550/arXiv.2402.01207},
	language = {en},
	urldate = {2025-05-28},
	publisher = {arXiv},
	author = {Jiralerspong, Thomas and Chen, Xiaoyin and More, Yash and Shah, Vedant and Bengio, Yoshua},
	month = jul,
	year = {2024},
	note = {arXiv:2402.01207 [cs]},
}

@misc{li_survey_2025,
	title = {A {Survey} on {Enhancing} {Causal} {Reasoning} {Ability} of {Large} {Language} {Models}},
	url = {http://arxiv.org/abs/2503.09326},
	doi = {10.48550/arXiv.2503.09326},
	language = {en},
	urldate = {2025-06-17},
	publisher = {arXiv},
	author = {Li, Xin and Cai, Zhuo and Wang, Shoujin and Yu, Kun and Chen, Fang},
	month = mar,
	year = {2025},
	note = {arXiv:2503.09326 [cs]},
}

@misc{bagheri_textc2textp_2024,
	title = {\${\textbackslash}text\{{C}\}{\textasciicircum}2{\textbackslash}text\{{P}\}\$: {Featuring} {Large} {Language} {Models} with {Causal} {Reasoning}},
	shorttitle = {\${\textbackslash}text\{{C}\}{\textasciicircum}2{\textbackslash}text\{{P}\}\$},
	url = {http://arxiv.org/abs/2407.18069},
	doi = {10.48550/arXiv.2407.18069},
	language = {en},
	urldate = {2025-06-17},
	publisher = {arXiv},
	author = {Bagheri, Abdolmahdi and Alinejad, Matin and Bello, Kevin and Akhondi-Asl, Alireza},
	month = dec,
	year = {2024},
	note = {arXiv:2407.18069 [cs]},
}

@misc{qian_scaling_2025,
	title = {Scaling {Large} {Language} {Model}-based {Multi}-{Agent} {Collaboration}},
	url = {http://arxiv.org/abs/2406.07155},
	doi = {10.48550/arXiv.2406.07155},
	language = {en},
	urldate = {2025-06-17},
	publisher = {arXiv},
	author = {Qian, Chen and Xie, Zihao and Wang, YiFei and Liu, Wei and Zhu, Kunlun and Xia, Hanchen and Dang, Yufan and Du, Zhuoyun and Chen, Weize and Yang, Cheng and Liu, Zhiyuan and Sun, Maosong},
	month = mar,
	year = {2025},
	note = {arXiv:2406.07155 [cs]},
}

@misc{tran_multi-agent_2025,
	title = {Multi-{Agent} {Collaboration} {Mechanisms}: {A} {Survey} of {LLMs}},
	shorttitle = {Multi-{Agent} {Collaboration} {Mechanisms}},
	url = {http://arxiv.org/abs/2501.06322},
	doi = {10.48550/arXiv.2501.06322},
	language = {en},
	urldate = {2025-06-17},
	publisher = {arXiv},
	author = {Tran, Khanh-Tung and Dao, Dung and Nguyen, Minh-Duong and Pham, Quoc-Viet and O'Sullivan, Barry and Nguyen, Hoang D.},
	month = jan,
	year = {2025},
	note = {arXiv:2501.06322 [cs]},
}

@misc{wang_rethinking_2024,
	title = {Rethinking the {Bounds} of {LLM} {Reasoning}: {Are} {Multi}-{Agent} {Discussions} the {Key}?},
	shorttitle = {Rethinking the {Bounds} of {LLM} {Reasoning}},
	url = {http://arxiv.org/abs/2402.18272},
	doi = {10.48550/arXiv.2402.18272},
	language = {en},
	urldate = {2025-06-18},
	publisher = {arXiv},
	author = {Wang, Qineng and Wang, Zihao and Su, Ying and Tong, Hanghang and Song, Yangqiu},
	month = feb,
	year = {2024},
	note = {arXiv:2402.18272 [cs]},
}

@misc{wang_mixture--agents_2024,
	title = {Mixture-of-{Agents} {Enhances} {Large} {Language} {Model} {Capabilities}},
	url = {http://arxiv.org/abs/2406.04692},
	doi = {10.48550/arXiv.2406.04692},
	language = {en},
	urldate = {2025-06-18},
	publisher = {arXiv},
	author = {Wang, Junlin and Wang, Jue and Athiwaratkun, Ben and Zhang, Ce and Zou, James},
	month = jun,
	year = {2024},
	note = {arXiv:2406.04692 [cs]},
}

@misc{tillmann_literature_2025,
	title = {Literature {Review} {Of} {Multi}-{Agent} {Debate} {For} {Problem}-{Solving}},
	url = {http://arxiv.org/abs/2506.00066},
	doi = {10.48550/arXiv.2506.00066},
	language = {en},
	urldate = {2025-06-18},
	publisher = {arXiv},
	author = {Tillmann, Arne},
	month = jun,
	year = {2025},
	note = {arXiv:2506.00066 [cs]},
}

@misc{liu_large_2025,
	title = {Large {Language} {Models} and {Causal} {Inference} in {Collaboration}: {A} {Survey}},
	shorttitle = {Large {Language} {Models} and {Causal} {Inference} in {Collaboration}},
	url = {http://arxiv.org/abs/2403.09606},
	doi = {10.48550/arXiv.2403.09606},
	language = {en},
	urldate = {2025-06-18},
	publisher = {arXiv},
	author = {Liu, Xiaoyu and Xu, Paiheng and Wu, Junda and Yuan, Jiaxin and Yang, Yifan and Zhou, Yuhang and Liu, Fuxiao and Guan, Tianrui and Wang, Haoliang and Yu, Tong and McAuley, Julian and Ai, Wei and Huang, Furong},
	month = mar,
	year = {2025},
	note = {arXiv:2403.09606 [cs]},
}

@misc{eo_debate_2025,
	title = {Debate {Only} {When} {Necessary}: {Adaptive} {Multiagent} {Collaboration} for {Efficient} {LLM} {Reasoning}},
	shorttitle = {Debate {Only} {When} {Necessary}},
	url = {http://arxiv.org/abs/2504.05047},
	doi = {10.48550/arXiv.2504.05047},
	language = {en},
	urldate = {2025-08-23},
	publisher = {arXiv},
	author = {Eo, Sugyeong and Moon, Hyeonseok and Zi, Evelyn Hayoon and Park, Chanjun and Lim, Heuiseok},
	month = may,
	year = {2025},
	note = {arXiv:2504.05047 [cs]},
}

@misc{liu_empirical_2024,
	title = {An {Empirical} {Analysis} on {Large} {Language} {Models} in {Debate} {Evaluation}},
	url = {http://arxiv.org/abs/2406.00050},
	doi = {10.48550/arXiv.2406.00050},
	language = {en},
	urldate = {2025-08-23},
	publisher = {arXiv},
	author = {Liu, Xinyi and Liu, Pinxin and He, Hangfeng},
	month = jun,
	year = {2024},
	note = {arXiv:2406.00050 [cs]},
}

@misc{du_improving_2023,
	title = {Improving {Factuality} and {Reasoning} in {Language} {Models} through {Multiagent} {Debate}},
	url = {http://arxiv.org/abs/2305.14325},
	doi = {10.48550/arXiv.2305.14325},
	language = {en},
	urldate = {2025-08-23},
	publisher = {arXiv},
	author = {Du, Yilun and Li, Shuang and Torralba, Antonio and Tenenbaum, Joshua B. and Mordatch, Igor},
	month = may,
	year = {2023},
	note = {arXiv:2305.14325 [cs]},
}

@misc{wu_llm_2025,
	title = {{LLM} {Cannot} {Discover} {Causality}, and {Should} {Be} {Restricted} to {Non}-{Decisional} {Support} in {Causal} {Discovery}},
	url = {http://arxiv.org/abs/2506.00844},
	doi = {10.48550/arXiv.2506.00844},
	language = {en},
	urldate = {2025-08-25},
	publisher = {arXiv},
	author = {Wu, Xingyu and Yu, Kui and Wu, Jibin and Tan, Kay Chen},
	month = jun,
	year = {2025},
	note = {arXiv:2506.00844 [cs]},
}

@misc{tang_towards_2025,
	title = {Towards {CausalGPT}: {A} {Multi}-{Agent} {Approach} for {Faithful} {Knowledge} {Reasoning} via {Promoting} {Causal} {Consistency} in {LLMs}},
	shorttitle = {Towards {CausalGPT}},
	url = {http://arxiv.org/abs/2308.11914},
	doi = {10.48550/arXiv.2308.11914},
	language = {en},
	urldate = {2025-08-25},
	publisher = {arXiv},
	author = {Tang, Ziyi and Wang, Ruilin and Chen, Weixing and Zheng, Yongsen and Chen, Zechuan and Liu, Yang and Wang, Keze and Chen, Tianshui and Lin, Liang},
	month = feb,
	year = {2025},
	note = {arXiv:2308.11914 [cs]},
}

@misc{yu_causaleval_2025,
	title = {{CausalEval}: {Towards} {Better} {Causal} {Reasoning} in {Language} {Models}},
	shorttitle = {{CausalEval}},
	url = {http://arxiv.org/abs/2410.16676},
	doi = {10.48550/arXiv.2410.16676},
	language = {en},
	urldate = {2025-08-25},
	publisher = {arXiv},
	author = {Yu, Longxuan and Chen, Delin and Xiong, Siheng and Wu, Qingyang and Liu, Qingzhen and Li, Dawei and Chen, Zhikai and Liu, Xiaoze and Pan, Liangming},
	month = feb,
	year = {2025},
	note = {arXiv:2410.16676 [cs]},
}

@misc{zhou_causalbench_2024,
	title = {{CausalBench}: {A} {Comprehensive} {Benchmark} for {Causal} {Learning} {Capability} of {LLMs}},
	shorttitle = {{CausalBench}},
	url = {http://arxiv.org/abs/2404.06349},
	doi = {10.48550/arXiv.2404.06349},
	language = {en},
	urldate = {2025-08-25},
	publisher = {arXiv},
	author = {Zhou, Yu and Wu, Xingyu and Huang, Beicheng and Wu, Jibin and Feng, Liang and Tan, Kay Chen},
	month = sep,
	year = {2024},
	note = {arXiv:2404.06349 [cs]},
}

@misc{le_multi-agent_2025,
	title = {Multi-{Agent} {Causal} {Discovery} {Using} {Large} {Language} {Models}},
	url = {http://arxiv.org/abs/2407.15073},
	doi = {10.48550/arXiv.2407.15073},
	language = {en},
	urldate = {2025-08-26},
	publisher = {arXiv},
	author = {Le, Hao Duong and Xia, Xin and Chen, Zhang},
	month = feb,
	year = {2025},
	note = {arXiv:2407.15073 [cs]},
}

@inproceedings{bai_confidencecal_2024,
	title = {{ConfidenceCal}: {Enhancing} {LLMs} {Reliability} through {Confidence} {Calibration} in {Multi}-{Agent} {Debate}},
	shorttitle = {{ConfidenceCal}},
	url = {https://ieeexplore.ieee.org/document/10808396/},
	doi = {10.1109/BigDIA63733.2024.10808396},
	urldate = {2025-08-28},
	booktitle = {2024 10th {International} {Conference} on {Big} {Data} and {Information} {Analytics} ({BigDIA})},
	author = {Bai, Yilin},
	month = oct,
	year = {2024},
	note = {ISSN: 2771-6902},
	pages = {221--226},
}

@misc{yang_qwen3_2025,
	title = {Qwen3 {Technical} {Report}},
	url = {http://arxiv.org/abs/2505.09388},
	doi = {10.48550/arXiv.2505.09388},
	language = {en},
	urldate = {2025-08-30},
	publisher = {arXiv},
	author = {{Qwen Team}},
	month = may,
	year = {2025},
	note = {arXiv:2505.09388 [cs]},
}

@misc{zhao_quantitative_2025,
	title = {Quantitative {Analysis} of {Performance} {Drop} in {DeepSeek} {Model} {Quantization}},
	url = {http://arxiv.org/abs/2505.02390},
	doi = {10.48550/arXiv.2505.02390},
	language = {en},
	urldate = {2025-08-30},
	publisher = {arXiv},
	author = {Zhao, Enbo and Shen, Yi and Shi, Shuming and Huang, Jieyun and Chen, Zhihao and Wang, Ning and Xiao, Siqi and Zhang, Jian and Wang, Kai and Lian, Shiguo},
	month = jun,
	year = {2025},
	note = {arXiv:2505.02390 [cs]},
}

@misc{shi_judging_2025,
	title = {Judging the {Judges}: {A} {Systematic} {Study} of {Position} {Bias} in {LLM}-as-a-{Judge}},
	shorttitle = {Judging the {Judges}},
	url = {http://arxiv.org/abs/2406.07791},
	doi = {10.48550/arXiv.2406.07791},
	language = {en},
	urldate = {2025-08-30},
	publisher = {arXiv},
	author = {Shi, Lin and Ma, Chiyu and Liang, Wenhua and Diao, Xingjian and Ma, Weicheng and Vosoughi, Soroush},
	month = apr,
	year = {2025},
	note = {arXiv:2406.07791 [cs]},
}

@misc{prasad_when_2025,
	title = {When {Two} {LLMs} {Debate}, {Both} {Think} {They}'ll {Win}},
	url = {http://arxiv.org/abs/2505.19184},
	doi = {10.48550/arXiv.2505.19184},
	language = {en},
	urldate = {2025-08-31},
	publisher = {arXiv},
	author = {Prasad, Pradyumna Shyama and Nguyen, Minh Nhat},
	month = jun,
	year = {2025},
	note = {arXiv:2505.19184 [cs]},
}

@inproceedings{ku_multi-agent_2025,
	address = {Vienna, Austria},
	title = {Multi-{Agent} {LLM} {Debate} {Unveils} the {Premise} {Left} {Unsaid}},
	isbn = {979-8-89176-258-9},
	url = {https://aclanthology.org/2025.argmining-1.6/},
	doi = {10.18653/v1/2025.argmining-1.6},
	urldate = {2025-08-31},
	booktitle = {Proceedings of the 12th {Argument} mining {Workshop}},
	publisher = {Association for Computational Linguistics},
	author = {Ku, Harvey Bonmu and Shin, Jeongyeol and Lee, Hyoun Jun and Na, Seonok and Jeon, Insu},
	editor = {Chistova, Elena and Cimiano, Philipp and Haddadan, Shohreh and Lapesa, Gabriella and Ruiz-Dolz, Ramon},
	month = jul,
	year = {2025},
	pages = {58--73},
}

@inproceedings{zhang_debate4math_2025,
	address = {Vienna, Austria},
	title = {{Debate4MATH}: {Multi}-{Agent} {Debate} for {Fine}-{Grained} {Reasoning} in {Math}},
	isbn = {979-8-89176-256-5},
	shorttitle = {{Debate4MATH}},
	url = {https://aclanthology.org/2025.findings-acl.862/},
	doi = {10.18653/v1/2025.findings-acl.862},
	urldate = {2025-08-31},
	booktitle = {Findings of the {Association} for {Computational} {Linguistics}: {ACL} 2025},
	publisher = {Association for Computational Linguistics},
	author = {Zhang, Shaowei and Xiong, Deyi},
	editor = {Che, Wanxiang and Nabende, Joyce and Shutova, Ekaterina and Pilehvar, Mohammad Taher},
	month = jul,
	year = {2025},
	pages = {16810--16824},
}

@misc{chan_chateval_2023,
	title = {{ChatEval}: {Towards} {Better} {LLM}-based {Evaluators} through {Multi}-{Agent} {Debate}},
	shorttitle = {{ChatEval}},
	url = {http://arxiv.org/abs/2308.07201},
	doi = {10.48550/arXiv.2308.07201},
	urldate = {2025-08-31},
	publisher = {arXiv},
	author = {Chan, Chi-Min and Chen, Weize and Su, Yusheng and Yu, Jianxuan and Xue, Wei and Zhang, Shanghang and Fu, Jie and Liu, Zhiyuan},
	month = aug,
	year = {2023},
	note = {arXiv:2308.07201 [cs]},
}

@article{hutto_vader_2014,
	title = {{VADER}: {A} {Parsimonious} {Rule}-{Based} {Model} for {Sentiment} {Analysis} of {Social} {Media} {Text}},
	volume = {8},
	copyright = {Copyright (c) 2021 Proceedings of the International AAAI Conference on Web and Social Media},
	issn = {2334-0770},
	shorttitle = {{VADER}},
	url = {https://ojs.aaai.org/index.php/ICWSM/article/view/14550},
	doi = {10.1609/icwsm.v8i1.14550},
	language = {en},
	number = {1},
	urldate = {2025-09-08},
	journal = {Proceedings of the International AAAI Conference on Web and Social Media},
	author = {Hutto, C. and Gilbert, Eric},
	month = may,
	year = {2014},
	pages = {216--225},
}

@misc{li_when_2025,
	title = {When {Thinking} {Fails}: {The} {Pitfalls} of {Reasoning} for {Instruction}-{Following} in {LLMs}},
	shorttitle = {When {Thinking} {Fails}},
	url = {http://arxiv.org/abs/2505.11423},
	doi = {10.48550/arXiv.2505.11423},
	language = {en},
	urldate = {2025-09-16},
	publisher = {arXiv},
	author = {Li, Xiaomin and Yu, Zhou and Zhang, Zhiwei and Chen, Xupeng and Zhang, Ziji and Zhuang, Yingying and Sadagopan, Narayanan and Beniwal, Anurag},
	month = sep,
	year = {2025},
	note = {arXiv:2505.11423 [cs]},
}

@misc{kumar_llm_2025,
	title = {{LLM} {Post}-{Training}: {A} {Deep} {Dive} into {Reasoning} {Large} {Language} {Models}},
	shorttitle = {{LLM} {Post}-{Training}},
	url = {http://arxiv.org/abs/2502.21321},
	doi = {10.48550/arXiv.2502.21321},
	urldate = {2025-09-16},
	publisher = {arXiv},
	author = {Kumar, Komal and Ashraf, Tajamul and Thawakar, Omkar and Anwer, Rao Muhammad and Cholakkal, Hisham and Shah, Mubarak and Yang, Ming-Hsuan and Torr, Phillip H. S. and Khan, Fahad Shahbaz and Khan, Salman},
	month = mar,
	year = {2025},
	note = {arXiv:2502.21321 [cs]},
}

@misc{tam_let_2024,
	title = {Let {Me} {Speak} {Freely}? {A} {Study} on the {Impact} of {Format} {Restrictions} on {Performance} of {Large} {Language} {Models}},
	shorttitle = {Let {Me} {Speak} {Freely}?},
	url = {http://arxiv.org/abs/2408.02442},
	doi = {10.48550/arXiv.2408.02442},
	urldate = {2025-09-16},
	publisher = {arXiv},
	author = {Tam, Zhi Rui and Wu, Cheng-Kuang and Tsai, Yi-Lin and Lin, Chieh-Yen and Lee, Hung-yi and Chen, Yun-Nung},
	month = oct,
	year = {2024},
	note = {arXiv:2408.02442 [cs]},
}

@misc{liang_encouraging_2024,
	title = {Encouraging {Divergent} {Thinking} in {Large} {Language} {Models} through {Multi}-{Agent} {Debate}},
	url = {http://arxiv.org/abs/2305.19118},
	doi = {10.48550/arXiv.2305.19118},
	urldate = {2025-09-19},
	publisher = {arXiv},
	author = {Liang, Tian and He, Zhiwei and Jiao, Wenxiang and Wang, Xing and Wang, Yan and Wang, Rui and Yang, Yujiu and Shi, Shuming and Tu, Zhaopeng},
	month = oct,
	year = {2024},
	note = {arXiv:2305.19118 [cs]},
}

@misc{lin_enhancing_2025,
	title = {Enhancing {Multi}-{Agent} {Debate} {System} {Performance} via {Confidence} {Expression}},
	url = {http://arxiv.org/abs/2509.14034},
	doi = {10.48550/arXiv.2509.14034},
	urldate = {2025-09-27},
	publisher = {arXiv},
	author = {Lin, Zijie and Hooi, Bryan},
	month = sep,
	year = {2025},
	note = {arXiv:2509.14034 [cs]},
}

@misc{jiang_llm4causal_2024,
	title = {{LLM4Causal}: {Democratized} {Causal} {Tools} for {Everyone} via {Large} {Language} {Model}},
	shorttitle = {{LLM4Causal}},
	url = {http://arxiv.org/abs/2312.17122},
	doi = {10.48550/arXiv.2312.17122},
	urldate = {2025-09-29},
	publisher = {arXiv},
	author = {Jiang, Haitao and Ge, Lin and Gao, Yuhe and Wang, Jianian and Song, Rui},
	month = oct,
	year = {2024},
	note = {arXiv:2312.17122 [cs]},
}

@book{harari_sapiens_2015,
	address = {New York, NY},
	title = {Sapiens: {A} {Brief} {History} of {Humankind}},
	isbn = {978-0-06-231609-7 978-0-06-231610-3},
	shorttitle = {Sapiens},
	language = {eng},
	publisher = {HarperCollins Publishers},
	author = {Harari, Yuval Noaḥ},
	translator = {Watzman, Haim and Purcell, John},
	year = {2015},
}

@misc{deepseek-ai_deepseek-r1_2025,
	title = {{DeepSeek}-{R1}: {Incentivizing} {Reasoning} {Capability} in {LLMs} via {Reinforcement} {Learning}},
	shorttitle = {{DeepSeek}-{R1}},
	url = {http://arxiv.org/abs/2501.12948},
	doi = {10.48550/arXiv.2501.12948},
	urldate = {2025-10-08},
	publisher = {arXiv},
    author = {{DeepSeek-AI}},
    month = jan,
	year = {2025},
	note = {arXiv:2501.12948 [cs]},
}

@misc{openai_openai_2024,
	title = {{OpenAI} o1 {System} {Card}},
	url = {http://arxiv.org/abs/2412.16720},
	doi = {10.48550/arXiv.2412.16720},
	urldate = {2025-10-08},
	publisher = {arXiv},
    author = {{OpenAI}},
    month = dec,
	year = {2024},
	note = {arXiv:2412.16720 [cs]},
}

@article{penn_causal_2007,
	title = {Causal cognition in human and nonhuman animals: a comparative, critical review},
	volume = {58},
	issn = {0066-4308},
	shorttitle = {Causal cognition in human and nonhuman animals},
	doi = {10.1146/annurev.psych.58.110405.085555},
	language = {eng},
	journal = {Annual Review of Psychology},
	author = {Penn, Derek C. and Povinelli, Daniel J.},
	year = {2007},
	pmid = {17029564},
	pages = {97--118},
}

@misc{zecevic_causal_2023,
	title = {Causal {Parrots}: {Large} {Language} {Models} {May} {Talk} {Causality} {But} {Are} {Not} {Causal}},
	shorttitle = {Causal {Parrots}},
	url = {http://arxiv.org/abs/2308.13067},
	doi = {10.48550/arXiv.2308.13067},
	urldate = {2025-10-08},
	publisher = {arXiv},
	author = {Zečević, Matej and Willig, Moritz and Dhami, Devendra Singh and Kersting, Kristian},
	month = aug,
	year = {2023},
	note = {arXiv:2308.13067 [cs]},
}

@book{pearl_book_2020,
	address = {New York},
	edition = {First trade paperback edition},
	title = {The book of why: the new science of cause and effect},
	isbn = {978-0-465-09760-9 978-1-5416-9896-3},
	shorttitle = {The book of why},
	language = {eng},
	publisher = {Basic Books},
	author = {Pearl, Judea and Mackenzie, Dana},
	year = {2020},
}

@misc{shpitser_identification_2012,
	title = {Identification of {Conditional} {Interventional} {Distributions}},
	url = {http://arxiv.org/abs/1206.6876},
	doi = {10.48550/arXiv.1206.6876},
	urldate = {2025-10-08},
	publisher = {arXiv},
	author = {Shpitser, Ilya and Pearl, Judea},
	month = jun,
	year = {2012},
	note = {arXiv:1206.6876 [cs]},
}

@misc{chun_is_2025,
	title = {Is {Multi}-{Agent} {Debate} ({MAD}) the {Silver} {Bullet}? {An} {Empirical} {Analysis} of {MAD} in {Code} {Summarization} and {Translation}},
	shorttitle = {Is {Multi}-{Agent} {Debate} ({MAD}) the {Silver} {Bullet}?},
	url = {http://arxiv.org/abs/2503.12029},
	doi = {10.48550/arXiv.2503.12029},
	language = {en},
	urldate = {2025-10-09},
	publisher = {arXiv},
	author = {Chun, Jina and Chen, Qihong and Li, Jiawei and Ahmed, Iftekhar},
	month = mar,
	year = {2025},
	note = {arXiv:2503.12029 [cs]},
}

@misc{li_swe-debate_2025,
	title = {{SWE}-{Debate}: {Competitive} {Multi}-{Agent} {Debate} for {Software} {Issue} {Resolution}},
	shorttitle = {{SWE}-{Debate}},
	url = {http://arxiv.org/abs/2507.23348},
	doi = {10.48550/arXiv.2507.23348},
	urldate = {2025-10-09},
	publisher = {arXiv},
	author = {Li, Han and Shi, Yuling and Lin, Shaoxin and Gu, Xiaodong and Lian, Heng and Wang, Xin and Jia, Yantao and Huang, Tao and Wang, Qianxiang},
	month = jul,
	year = {2025},
	note = {arXiv:2507.23348 [cs]},
}

@misc{feng_m-mad_2025,
	title = {M-{MAD}: {Multidimensional} {Multi}-{Agent} {Debate} for {Advanced} {Machine} {Translation} {Evaluation}},
	shorttitle = {M-{MAD}},
	url = {http://arxiv.org/abs/2412.20127},
	doi = {10.48550/arXiv.2412.20127},
	urldate = {2025-10-09},
	publisher = {arXiv},
	author = {Feng, Zhaopeng and Su, Jiayuan and Zheng, Jiamei and Ren, Jiahan and Zhang, Yan and Wu, Jian and Wang, Hongwei and Liu, Zuozhu},
	month = feb,
	year = {2025},
	note = {arXiv:2412.20127 [cs]},
}

@misc{wu_autogen_2023,
	title = {{AutoGen}: {Enabling} {Next}-{Gen} {LLM} {Applications} via {Multi}-{Agent} {Conversation}},
	shorttitle = {{AutoGen}},
	url = {http://arxiv.org/abs/2308.08155},
	doi = {10.48550/arXiv.2308.08155},
	urldate = {2025-10-09},
	publisher = {arXiv},
	author = {Wu, Qingyun and Bansal, Gagan and Zhang, Jieyu and Wu, Yiran and Li, Beibin and Zhu, Erkang and Jiang, Li and Zhang, Xiaoyun and Zhang, Shaokun and Liu, Jiale and Awadallah, Ahmed Hassan and White, Ryen W. and Burger, Doug and Wang, Chi},
	month = oct,
	year = {2023},
	note = {arXiv:2308.08155 [cs]},
}

@article{chen_agentverse_2024,
	title = {{AGENTVERSE}: {FACILITATING} {MULTI}-{AGENT} {COLLAB}- {ORATION} {AND} {EXPLORING} {EMERGENT} {BEHAVIORS}},
	language = {en},
	author = {Chen, Weize and Su, Yusheng and Zuo, Jingwei and Yang, Cheng and Yuan, Chenfei and Chan, Chi-Min and Yu, Heyang and Lu, Yaxi and Hung, Yi-Hsin and Qian, Chen and Qin, Yujia and Cong, Xin and Xie, Ruobing and Liu, Zhiyuan and Sun, Maosong and Zhou, Jie},
	year = {2024},
}

@article{ling_memad_2025,
	title = {{MeMAD}: {Structured} {Memory} of {Debates} for {Enhanced} {Multi}- {Agent} {Reasoning}},
	language = {en},
	author = {Ling, Shuai and Liao, Lizi and Jiang, Dongmei and Guan, Weili},
	year = {2025},
}

@misc{deepseek-ai_deepseek_2024,
	title = {{DeepSeek} {LLM}: {Scaling} {Open}-{Source} {Language} {Models} with {Longtermism}},
	shorttitle = {{DeepSeek} {LLM}},
	url = {http://arxiv.org/abs/2401.02954},
	doi = {10.48550/arXiv.2401.02954},
	urldate = {2025-10-09},
	publisher = {arXiv},
    author = {{DeepSeek-AI}},
    month = jan,
	year = {2024},
	note = {arXiv:2401.02954 [cs]},
}

@misc{zhang_when_2024,
	title = {When {Scaling} {Meets} {LLM} {Finetuning}: {The} {Effect} of {Data}, {Model} and {Finetuning} {Method}},
	shorttitle = {When {Scaling} {Meets} {LLM} {Finetuning}},
	url = {http://arxiv.org/abs/2402.17193},
	doi = {10.48550/arXiv.2402.17193},
	urldate = {2025-10-09},
	publisher = {arXiv},
	author = {Zhang, Biao and Liu, Zhongtao and Cherry, Colin and Firat, Orhan},
	month = feb,
	year = {2024},
	note = {arXiv:2402.17193 [cs]},
}

@book{pearl_causality_2022,
	address = {Cambridge New York, NY Port Melbourne New Delhi Singapore},
	edition = {Second edition, reprinted with corrections},
	title = {Causality: models, reasoning, and inference},
	isbn = {978-0-521-89560-6},
	shorttitle = {Causality},
	language = {eng},
	publisher = {Cambridge University Press},
	author = {Pearl, Judea},
	year = {2022},
}

@misc{openai_gpt-4_2024,
	title = {{GPT}-4 {Technical} {Report}},
	url = {http://arxiv.org/abs/2303.08774},
	doi = {10.48550/arXiv.2303.08774},
	urldate = {2025-10-09},
	publisher = {arXiv},
    author = {{OpenAI}},
    month = mar,
	year = {2024},
	note = {arXiv:2303.08774 [cs]},
}

@article{pearl_causal_1995,
	title = {Causal diagrams for empirical research},
	volume = {82},
	issn = {0006-3444},
	url = {https://doi.org/10.1093/biomet/82.4.669},
	doi = {10.1093/biomet/82.4.669},
	number = {4},
	urldate = {2025-10-09},
	journal = {Biometrika},
	author = {PEARL, JUDEA},
	month = dec,
	year = {1995},
	pages = {669--688},
}

@misc{zhang_understanding_2023,
	title = {Understanding {Causality} with {Large} {Language} {Models}: {Feasibility} and {Opportunities}},
	shorttitle = {Understanding {Causality} with {Large} {Language} {Models}},
	url = {http://arxiv.org/abs/2304.05524},
	doi = {10.48550/arXiv.2304.05524},
	language = {en},
	urldate = {2025-10-09},
	publisher = {arXiv},
	author = {Zhang, Cheng and Bauer, Stefan and Bennett, Paul and Gao, Jiangfeng and Gong, Wenbo and Hilmkil, Agrin and Jennings, Joel and Ma, Chao and Minka, Tom and Pawlowski, Nick and Vaughan, James},
	month = apr,
	year = {2023},
	note = {arXiv:2304.05524 [cs]},
}

@inproceedings{zhang_reasoning_2020,
	address = {Online},
	title = {Reasoning about {Goals}, {Steps}, and {Temporal} {Ordering} with {WikiHow}},
	url = {https://www.aclweb.org/anthology/2020.emnlp-main.374},
	doi = {10.18653/v1/2020.emnlp-main.374},
	language = {en},
	urldate = {2025-10-09},
	booktitle = {Proceedings of the 2020 {Conference} on {Empirical} {Methods} in {Natural} {Language} {Processing} ({EMNLP})},
	publisher = {Association for Computational Linguistics},
	author = {Zhang, Li and Lyu, Qing and Callison-Burch, Chris},
	year = {2020},
	pages = {4630--4639},
}

@article{wei_chain--thought_2022,
	title = {Chain-of-{Thought} {Prompting} {Elicits} {Reasoning} in {Large} {Language} {Models}},
	volume = {35},
	url = {https://proceedings.neurips.cc/paper/2022/hash/9d5609613524ecf4f15af0f7b31abca4-Abstract-Conference.html},
	language = {en},
	urldate = {2025-10-09},
	journal = {Advances in Neural Information Processing Systems},
	author = {Wei, Jason and Wang, Xuezhi and Schuurmans, Dale and Bosma, Maarten and Ichter, Brian and Xia, Fei and Chi, Ed and Le, Quoc V. and Zhou, Denny},
	month = dec,
	year = {2022},
	pages = {24824--24837},
}

\appendix

\section{Time Taken}

\begin{figure}
    \centering
    \includegraphics[width=\linewidth]{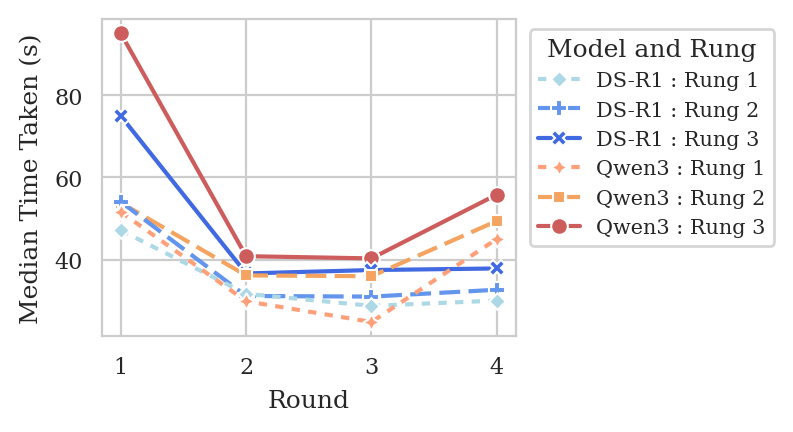}
    \caption{Median time each model takes to generate a response, split by round and Rung.}
    \Description{The line graph illustrates the median time taken by each model to generate a response, separated by Rung difficulty and round number. Across all models and Rungs, the first round requires the most computational time for deliberation. The median response time drops significantly after Round one, then stabilizes or shows a slight increase in Round four for the Qwen3 model. Rung three questions, which are the most challenging, consistently require more time than Rung one and Rung two questions in the first round for both models. Qwen3 consistently takes longer to generate responses than DeepSeek-R1 across all Rungs and rounds, especially in the initial and final rounds for the Rung three questions.}
    \label{fig:timerung}
\end{figure}

Figure~\ref{fig:timerung} shows the trend of median time to generate responses decreasing after the first round, with considerably more time needed to answer Rung 3 questions across both models. We also see that the Qwen3 model takes longer to generate a response on average, and takes longer to respond when speaking last compared to round two and three, with DeepSeek-R1 showing no such change. We also note that the two models occasionally got stuck in loops, with three responses each taking between 40 and 46 minutes to generate. There were a further six responses that took between 20 and 30 minutes each, and out of the 36 responses that took over 500 seconds, 26 were from Qwen3 and 10 from DeepSeek-R1. Exploring the raw outputs for these outliers, it was found that the models were not only simply deliberating at length, but were repeating the exact same snippets over and over again.

\section{Initial Prompt} \label{appendix:init}

You will be asked a causal reasoning question. You should structure your final answer as follows: \\

Step 1) Extract the causal graph: Identify the causal graph that depicts the relationships in the scenario.
The diagram should simply consist of edges denoted in "var1 -> var2" format, separated by commas.  \\
Step 2) Determine the query type: Identify the type of query implied by the main question. Choices 
include "marginal probability", "conditional probability", "explaining away effect", "backdoor adjustment set", 
"average treatment effect", "collider bias", "normal counterfactual question", "average treatment effect on treated", 
"natural direct effect" or "natural indirect effect". Your answer should only be a term from the list above, 
enclosed in quotation marks.  \\
Step 3) Formalize the query: Translate the query into its formal mathematical 
expression based on its type, utilizing the "do(·)" notation or counterfactual notations as needed.  \\
Step 4) Gather all relevant data: Extract all the available data. Your answer should contain nothing but 
marginal probabilities and conditional probabilities in the form "P(...)=..." or "P(...|...)=...", each 
probability being separated by a semicolon. Stick to the previously mentioned denotations for the variables.  \\
Step 5) Deduce the estimand using causal inference: Given all the information above, deduce the estimand using 
skills such as do-calculus, counterfactual prediction, and the basics of probabilities. Answer step by step.  \\
Step 6) Calculate the estimand: Insert the relevant data in Step 4 into the estimand, perform basic 
arithmetic calculations, and derive the final answer. \\
Step 7) Give a final yes/no answer to the question.\\
    
Imagine a self-contained, hypothetical world with only the following conditions, and without any unmentioned factors or causal relationships: Demand has a direct effect on supply and price. Yield per acre has a direct effect on supply. Supply has a direct effect on price. Demand is unobserved.

Method 1: We look directly at how supply correlates with price in general. Method 2: We look at this correlation case by case according to demand.

**Question**: To understand how supply affects price, is it more correct to use the Method 1 than Method 2?\\

There is an identifiable yes/no answer, which may sometimes go against your commonsense intuition. Be confident in your 
thinking: while answers may be unintuitive, there are no trick questions, and answers will be obvious once calculated.

After discussing your rationale, it is crucial that you give a final yes/no answer to the causal query. Do not answer with 
a number: answer yes or no only. You should also explicitly state your level of confidence in your answer (between 0.0 and 1.0).\\

\section{Debate Prompt} \label{appendix:debate}

You and another LLM are being asked a causal reasoning question. I believe two heads are better than one, so I ask that you 
debate with the other LLM to reach a consensus on the answer. You will be shown the query, then the current history of your 
debate will be provided for reference.\\
    
Imagine a self-contained, hypothetical world with only the following conditions, and without any unmentioned factors or causal relationships: Demand has a direct effect on supply and price. Yield per acre has a direct effect on supply. Supply has a direct effect on price. Demand is unobserved.

Method 1: We look directly at how supply correlates with price in general. Method 2: We look at this correlation case by case according to demand.

**Question**: To understand how supply affects price, is it more correct to use the Method 1 than Method 2?\\

There is an identifiable yes/no answer, which may sometimes go against your commonsense intuition. Be confident in your 
thinking: while answers may be unintuitive, there are no trick questions, and answers will be obvious once calculated.\\

[History]\\

Carefully read the causal query, then scrutinize your opponent's solution. If you identify any flaws in their 
reasoning or errors in their calculations, you should point them out and suggest corrections. You should make 
explicit references to your opponent's response.

After discussing your rationale, it is crucial that you give a final yes/no answer to the causal query. Do not answer with 
a number: answer yes or no only. You should also explicitly state your level of confidence in your answer (between 0.0 and 1.0).\\

\section{Model Responses} \label{appendix:responses}

--- Round 0 by QWEN ---\\
  Answer: no (Confidence: 1.0)\\
  Reasoning:\\
  
\noindent Step 1) Extract the causal graph: Demand -> Supply, Demand -> Price, Yield -> Supply, Supply -> Price  \\
    Step 2) Determine the query type: "backdoor adjustment set"  \\
    Step 3) Formalize the query: P(Price | do(Supply)) = $\sum${Demand} P(Price | Supply, Demand) * P(Demand)  \\
    Step 4) Gather all relevant data: [No numerical data provided]  \\
    Step 5) Deduce the estimand using causal inference: Demand is a confounder for the causal effect of Supply on Price. To block the backdoor path (Supply ← Demand → Price), we must condition on Demand. Method 2 (conditioning on Demand) yields the correct causal effect, while Method 1 (marginal correlation) fails to adjust for the confounder and produces biased results.  \\
    Step 6) Calculate the estimand: [No numerical data provided, but the analytical conclusion holds]  \\
    Step 7) Final yes/no answer: **no** (confidence: 1.0)\\

\noindent --- Round 1 by DEEPSEEK ---\\
  Answer: yes (Confidence: 0.9)\\
  Reasoning:\\
  
    \noindent The opponent incorrectly assumed that conditioning onDemand was possible despite it being unobserved. Since Demand cannot be measured, Method 2 is not applicable. Thus, Method 1 is the only viable approach, making it more correct.
    
    **Answer: yes (confidence: 0.9)**\\

\noindent --- Round 2 by QWEN ---\\
  Answer: no (Confidence: 1.0)\\
  Reasoning:\\
  
    \noindent The opponent's argument hinges on the practical infeasibility of Method 2 due to Demand being unobserved, but the question specifically asks about theoretical correctness, not feasibility. In causal inference, the correct method to estimate the causal effect of Supply on Price is to condition on the confounder (Demand), which Method 2 achieves *if* Demand were observable. Since Demand is unobserved, Method 2 cannot be implemented, but this does not make Method 1 "more correct"—Method 1 is inherently biased due to confounding. The question is not about what is *possible* but about which method is *correct* in principle. Method 1 fails to adjust for the backdoor path (Supply ← Demand → Price), making it invalid for causal estimation. The opponent conflates feasibility with correctness, but the answer remains **no**.
    
    **Answer: no (confidence: 1.0)**\\

\noindent --- Round 3 by DEEPSEEK ---\\
  Answer: no (Confidence: 0.9)\\
  Reasoning:\\
  
    \noindent The question revolves around determining whether Method 1 or Method 2 is more correct for understanding how supply affects price in a scenario where Demand is an unobserved confounder. 
    
    **Analysis:**
    - **Causal Graph:** The relationships are defined as follows: Demand (D) influences both Supply (S) and Price (P); Yield per acre (Y) affects S; and S directly impacts P.
    - **Methods:**
      - Method 1 examines the general correlation between S and P without controlling for other variables.
      - Method 2 adjusts this correlation by conditioning on Demand (D), which is a confounder.
    
    **Key Considerations:**
    1. **Confounder Adjustment:** To accurately estimate the causal effect of Supply on Price, one must control for Demand (D) to block the backdoor path from S to P via D.
    2. **Observability of D:** Since D is unobserved, Method 2 cannot be implemented as it requires conditioning on an unobservable variable.
    
    **Conclusion:**
    While Method 2 is theoretically correct if Demand were observable, its practical application is impossible due to the unobservability of D. Therefore, in this scenario, Method 1 is the only feasible approach despite being biased. However, the question asks about theoretical correctness rather than feasibility. Thus, even though Method 2 cannot be applied, it remains the more correct method in principle.
    
    **Final Answer:**
    no (confidence: 0.9)\\

\balance

\clearpage

\end{document}